\documentclass[sigconf]{acmart}
\AtBeginDocument{%
  }

\copyrightyear{2026}
\acmYear{2026}
\setcopyright{cc}
\setcctype{by}
\acmConference[MM '26] {Proceedings of the 34th ACM International Conference on Multimedia}{November 10--14, 2026}{Rio de Janeiro, Brazil.}
\acmBooktitle{Proceedings of the 34th ACM International Conference on Multimedia (MM '26), November 10--14, 2026, Rio de Janeiro, Brazil}
\acmISBN{979-8-4007-2213-4/2026/11}
\acmDOI{10.1145/3767308.3835750}

\usepackage{amsmath,amsfonts}
\usepackage{algorithmic}
\usepackage{algorithm}
\usepackage{array}
\usepackage{textcomp}
\usepackage{stfloats}
\usepackage{url}
\usepackage{verbatim}
\usepackage{graphicx}
\usepackage{subcaption}
\usepackage{tcolorbox}
\usepackage{float}
\usepackage{colortbl}
\usepackage{xcolor}  
\usepackage{multirow}
\usepackage{xspace}
\usepackage{enumitem}
\usepackage[normalem]{ulem}

\begin{document}

\title{Diagnose-to-Generate: Eliminating Structural Distortions in Flow Models via Patch-Level Grounded Alignment}

\author{Yifan Li}
\orcid{0009-0004-6097-2271}
\affiliation{
  \institution{University of Science and Technology of China}
  \city{Hefei}
  \state{AnHui Province}
  \country{China}
}
\additionalaffiliation{
  \institution{Meituan Inc.}
  \city{Beijing}
  \country{China}
}
  \email{liyifancqu@163.com}

\author{Lishuai Gao}
\orcid{0009-0000-1931-2889}
\correspondingauthor
\affiliation{
  \institution{Meituan Inc.}
  \city{Beijing}
   \country{China}
}
  \email{gaolishuai0425@163.com}

\author{Tong Zhang}
\orcid{0009-0003-4877-3934}
\affiliation{
  \institution{Meituan Inc.}
  \city{Beijing}
  \country{China}
}
  \email{zhangtong65@meituan.com}

\author{Qilong Huang}
\orcid{0009-0008-4676-1784}
\affiliation{
  \institution{Meituan Inc.}
  \city{Beijing}
  \country{China}
}
  \email{huangqilonghql@gmail.com}

\author{Chen Zhao}
\orcid{0009-0000-3762-3728}
\affiliation{
  \institution{Meituan Inc.}
  \city{Beijing}
  \country{China}
}
  \email{2518628273@qq.com}

\author{Zhuoliang Kang}
\orcid{0009-0000-3119-503X}
\affiliation{
  \institution{Meituan Inc.}
  \city{Beijing}
  \country{China}
}
  \email{kangzhuoliang@meituan.com}

\author{Jie Hu}
\orcid{0000-0002-5150-1003}
\affiliation{
  \institution{Meituan Inc.}
  \city{Beijing}
  \country{China}
}
  \email{hujiemr@163.com}

\author{Yujie Zhong}
\orcid{0000-0001-8383-3275}
\affiliation{
  \institution{Meituan Inc.}
  \city{Beijing}
  \country{China}
}
  \email{jaszhong@hotmail.com}

\author{Fenghe Tang}
\orcid{0009-0009-6193-4855}
\affiliation{
  \institution{University of Science and Technology of China}
  \city{Hefei}
  \state{AnHui Province}
  \country{China}
}
  \email{fhtan9@mail.ustc.edu.cn}

\author{S Kevin Zhou}
\orcid{0000-0002-6881-4444}
\affiliation{
  \institution{University of Science and Technology of China}
  \city{Hefei}
  \state{AnHui Province}
  \country{China}
}
  \email{s.kevin.zhou@gmail.com}

\author{Xiaoming Wei}
\orcid{0000-0002-7471-8344}
\affiliation{
  \institution{Meituan Inc.}
  \city{Beijing}
  \country{China}
}
  \email{weixiaoming@meituan.com}

\renewcommand{\shortauthors}{Lishuai Gao et al.}

\begin{abstract}
Flow-based image generators achieve strong semantic synthesis but still produce localized structural distortions, such as malformed hands and limbs. Existing post-training methods typically optimize global scalar rewards, which provide weak spatial credit assignment and may alter well-formed regions while correcting local defects. We present \textit{Diagnose-to-Generate (D2G)}, a grounded alignment framework that separates visual diagnosis from generative optimization. We construct \textbf{Achilles-600K}, a region-annotated dataset for structural anomalies, and train \textbf{ARGUS} through supervised cold-start and Region-Aware Group Relative Policy Optimization to predict anomaly categories and bounding boxes. These diagnoses are converted into confidence-weighted spatial rewards for our \textbf{Patch-level Flow-GRPO}. We further identify that global stochastic exploration can shift semantic regions across group rollouts, making patch-wise relative advantages spatially inconsistent. Our \textbf{Trajectory-Preserving Stochastic Sampling (TPSS)} anchors early denoising with deterministic ODE trajectories and introduces stochasticity only at later steps, thereby limiting intra-group spatial drift while retaining local exploration. Experiments on FLUX.1-dev and SD3.5 show improved structural fidelity with preserved FID and CLIP scores. Consistent gains under HumanRefiner, zero-shot GPT-4.1, and blinded human evaluation further demonstrate that the improvements extend beyond the training reward model.
\end{abstract}





\maketitle

\section{Introduction}
\label{intro}

Diffusion and flow-based models~\cite{flow_matching,ho2020denoising,liu2022flow,song2020score} generate images and videos with increasingly convincing composition, appearance, and prompt adherence~\cite{qwen_image,zimage,wan2025,deng2505emerging,gao2025seedance,kong2024hunyuanvideo}. Yet a plausible scene can still contain an extra finger, a malformed limb, or an object interaction that violates physical structure. These errors may occupy only a small fraction of an image but substantially affect its credibility and usability. Their persistence exposes a gap between global visual quality and local structural correctness: improving composition or aesthetics does not necessarily correct the geometry of individual entities.

\begin{figure}[t]
    \centering
    \includegraphics[width=\linewidth]{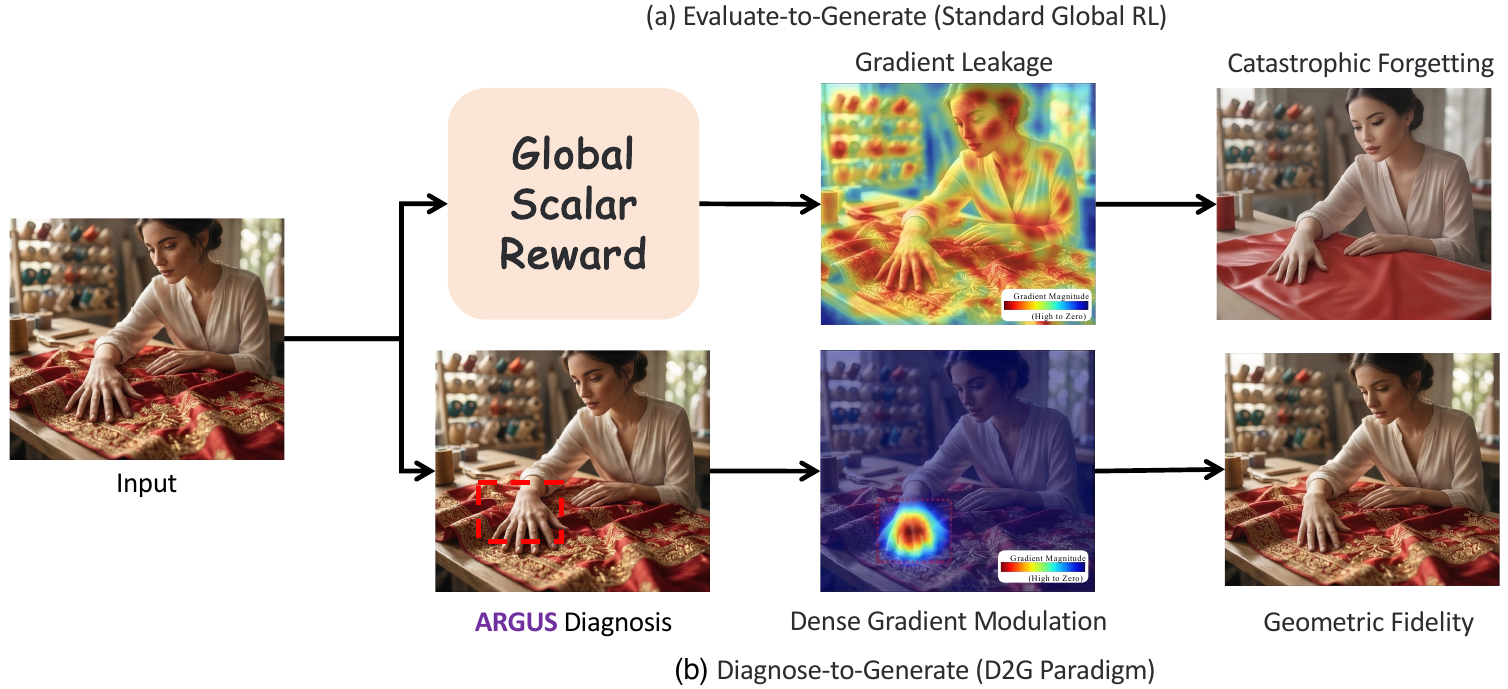}
    \caption{\textbf{Illustration of global and grounded structural alignment.} \textbf{(a)} An image-level reward does not identify the region responsible for a structural defect. \textbf{(b)} D2G uses ARGUS diagnoses to construct spatial rewards, concentrating structural feedback around anomalous regions while retaining a global quality objective.}
    \label{fig:paradigm_comparison}
\end{figure}

Reward-based optimization and preference learning provide practical routes to post-training alignment~\cite{ouyang2022training,rafailov2023direct,deepseekr1}. In visual generation, however, the feedback often takes the form of a scalar reward for an entire image. Such a reward can distinguish preferred outputs without identifying the regions responsible for that preference. A small anatomical defect may be outweighed by favorable aesthetics, while an unfavorable score does not distinguish the defective region from valid surrounding content. Applying this feedback throughout the spatial domain can therefore change well-formed content while attempting to correct a local error. Figure~\ref{fig:paradigm_comparison}(a) illustrates this \emph{spatial credit assignment} problem: effective structural alignment requires not only an assessment of image quality, but also evidence of where correction is needed.

Spatially resolved rewards introduce a further challenge when used for patch-wise group-relative optimization. Flow-GRPO and related methods introduce stochastic differential equation (SDE) sampling into deterministic flow trajectories to obtain online rollouts~\cite{liu2025flow,dancegrpo}. Early stochastic perturbations can change object positions and shapes across samples in the same group. A fixed latent coordinate may then correspond to a hand in one rollout and background in another. We refer to this loss of spatial correspondence as \emph{trajectory divergence}. Computing coordinate-wise relative advantages across these samples compares different semantic regions, making localized reward assignment less reliable even when each individual diagnosis is accurate.

We propose \textbf{Diagnose-to-Generate (D2G)}, which couples grounded structural diagnosis with a sampling procedure designed for patch-wise reward comparison (Figure~\ref{fig:paradigm_comparison}(b)). Its diagnostic foundation is \textbf{Achilles-600k}, a dataset of generated images with structural labels and localized anomaly annotations, accompanied by the separately curated \textbf{Achilles-Benchmark} for diagnostic and generative evaluation. We use human-verified labels and boxes to synthesize Reverse Chain-of-Thought (Reverse-CoT) explanations for a 10K-example cold-start subset. Conditioning explanations on verified defects reduces reliance on open-ended artifact discovery during data synthesis. We then train \textbf{ARGUS} (Anomaly Recognition and Grounded Understanding Specialist), first through supervised initialization and then through Region-Aware GRPO. Hungarian bipartite GIoU matching and a localization consistency gate connect category-level diagnostic rewards to spatial evidence, encouraging ARGUS to identify both the structural error and its location.

To transfer these diagnoses to the generator, we introduce \textbf{Patch-level Flow-GRPO} with \textbf{Trajectory-Preserving Stochastic Sampling (TPSS)}. TPSS retains deterministic ordinary differential equation (ODE) sampling during early denoising and restricts stochastic exploration to later steps, after the scene layout has been established. This reduces intra-group spatial drift and makes patch-wise comparisons more consistent without requiring identical final images. The frozen ARGUS model diagnoses each generated rollout, and its bounding boxes are mapped to latent tokens to construct spatial penalties. These penalties are combined with an image-level quality reward and normalized across the group at each spatial location. The resulting advantages concentrate structural optimization near diagnosed defects while retaining supervision for global image quality. D2G thus uses localized diagnosis to improve the generator itself, rather than treating diagnosis only as a post-generation assessment.

We evaluate ARGUS across diagnostic backbones and examine D2G on both in-distribution and unseen prompts. On FLUX.1-dev, D2G improves the Achilles structural fidelity score from $62.16$ to $85.60$, while FID-30K remains close to the base model ($18.35$ versus $18.42$). Because ARGUS also supplies the training reward, we complement its assessment with independent HumanRefiner~\cite{HumanRefiner}, zero-shot GPT-4.1, and double-blind human evaluation. HumanRefiner reports an anomaly-free rate of $74.8\%$, compared with $42.2\%$ for the base model, and the human study reports an $82.5\%$ preference win rate. These evaluations support structural improvements beyond the training evaluator, while ablations examine the separate contributions of grounded supervision, spatial reward assignment, and trajectory-preserving exploration.
\section{Related Work}

\subsection{Reward Models for Image Generation}

Image-generation evaluation has progressed from distribution-level metrics such as Inception Score and FID~\cite{IS,FID} to learned preference models that better reflect human judgments of visual quality and text--image alignment~\cite{kirstain2023pick,hpsv2,hpsv3,xu2023imagereward}. More recently, Multimodal Large Language Models (MLLMs) have been adapted as evaluators and reward models~\cite{Q-align,unifiedreward,Visualquality-r1,hpsv3,Qinsight}, providing richer semantic assessments and sometimes interpretable Chain-of-Thought feedback. Most methods, however, ultimately reduce an image to a global score or a small set of holistic attributes. Such feedback is effective for preference alignment but offers limited spatial attribution when the failure is confined to a small region, such as a malformed hand or an implausible object interaction.

Several approaches provide localized assessments. HADM~\cite{hadm} evaluates human anomalies through pixel-level comparison, while ViT-HD~\cite{ma2025evaluating} and HumanRefiner~\cite{HumanRefiner} specialize in human-centric structural defects. RichHF~\cite{RichHF} and HEIE~\cite{yang2025heie} produce spatial quality maps, offering finer feedback than scalar rewards. Reasoning-based evaluators including MagicAssessor~\cite{wang2025magicmirror} and REACT~\cite{wang2026thinking} improve interpretability through structured textual analysis. Nevertheless, these methods either focus on restricted domains or do not optimize bounding-box localization for downstream generator updates. ARGUS instead formulates structural assessment as grounded diagnosis: it jointly predicts anomaly categories, spatial coordinates, and diagnostic rationales across human, animal, interaction, and object-level distortions. Its outputs are convertible into localized rewards for generative alignment.

\subsection{RL for Generative Alignment}

Reinforcement learning has become an important mechanism for aligning foundation models with task objectives and human preferences~\cite{ouyang2022training,qwen3vl,deepseekr1,hong2025glm}. PPO~\cite{ppo} optimizes policies under an explicit reward model, while DPO~\cite{rafailov2023direct} replaces online policy optimization with preference-based supervision. GRPO~\cite{deepseekmath,deepseekr1} further removes the need for a learned critic by estimating advantages relative to samples within the same rollout group. This formulation has been adopted to improve reasoning in MLLMs~\cite{wang2025internvl3,shen2025vlm} and unified multimodal models~\cite{T2i-r1,Re-Align}.

For visual generation, recent work has explored both local preference alignment and online reinforcement learning. LocalDPO~\cite{localdpo}, RePlan~\cite{RePlan}, TextPecker~\cite{zhu2026textpecker}, and GIE-Bench~\cite{qian2025gie} study region-aware preferences, localized editing, or structural-error evaluation, highlighting the value of spatially resolved supervision. In parallel, Flow-GRPO~\cite{liu2025flow}, DanceGRPO~\cite{dancegrpo}, and MixGRPO~\cite{mixgrpo} extend GRPO to flow-based generators~\cite{flow_matching}. These methods introduce stochastic differential equation sampling~\cite{song2020score} into otherwise deterministic flow trajectories, enabling online exploration for image and video generation models~\cite{qwen_image,zimage,longcatvideo,spatialscore,RealGen}.

Existing flow-based RL methods compute a scalar reward for each rollout and apply stochastic exploration throughout the latent trajectory. This design is suitable for global objectives, but it becomes problematic for patch-level rewards: early stochastic perturbations can change object locations and shapes across samples in the same group, so a fixed spatial coordinate no longer refers to a consistent semantic region. Patch-level Flow-GRPO addresses this mismatch by preserving the early deterministic trajectory and restricting stochastic exploration to later denoising stages. The rollout group maintains stronger spatial correspondence, allowing ARGUS diagnoses to supervise localized regions without replacing the global semantic and aesthetic objectives.
\section{Achilles-600k and Achilles-Benchmark}
\label{sec:dataset}

\begin{figure*}[t]
    \centering
    \includegraphics[width=0.97\linewidth]{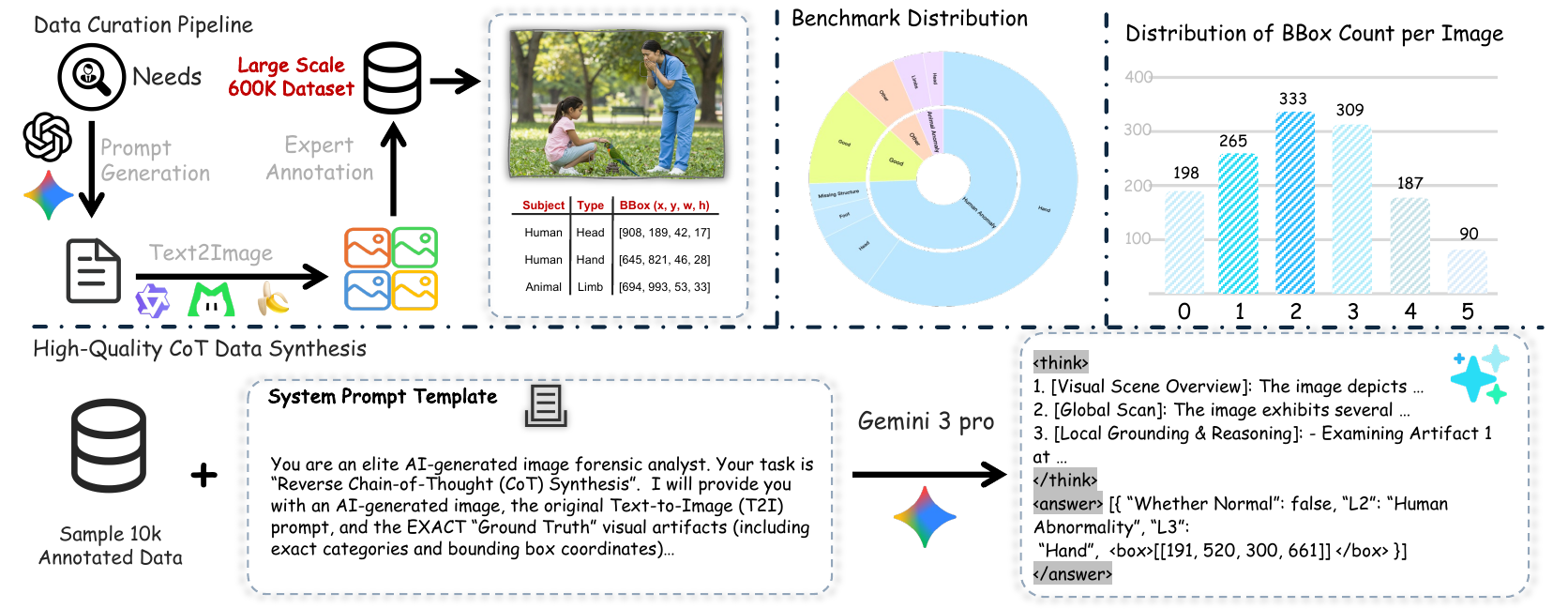}
    \caption{\textbf{Achilles-600k construction and benchmark statistics.}
    Diverse generators provide images for human-verified structural
    annotation, from which 10K grounded Reverse-CoT instances are selected
    for diagnostic cold-start training. The Achilles-Benchmark is independently
    curated and covers diverse anomaly categories, object scales, and numbers
    of localized defects.}
    \label{fig:dataset_pipeline}
    % \vspace{-2mm}
\end{figure*}

Existing image-alignment datasets primarily provide image-level
preferences or holistic quality labels~\cite{kirstain2023pick,hpsv3},
which are insufficient for learning where a localized structural failure
occurs. We therefore construct \textbf{Achilles-600k}, a region-grounded
dataset of generated images with anomaly categories and bounding-box
annotations, together with the independently curated
\textbf{Achilles-Benchmark} for diagnostic and generative evaluation.
Figure~\ref{fig:dataset_pipeline} summarizes the construction pipeline
and benchmark statistics.

\subsection{Dataset Construction}
\label{sec:data_construction}

\paragraph{Image collection and taxonomy.}
To reduce dependence on a single generator, source images are synthesized
using Z-image~\cite{zimage}, Qwen-Image~\cite{qwen_image},
LongCat-Image~\cite{LongCat-Image}, and Nano Banana
Pro~\cite{nano-banana-pro}. Their prompts are generated with GPT-4.1
and Gemini 3.1 Pro~\cite{GPT-4.1,gemini31pro} and vary in subject,
composition, environment, camera view, and visual style. The resulting
collection covers four top-level structural categories: abnormal
interactions, human anatomy, animal anatomy, and object morphology,
with additional fine-grained subcategories. Annotators assign category
labels and tight rectangular boxes to each localized defect; images
without identifiable structural anomalies are retained as normal
examples.

Achilles-600k contains 585K training images and 15K validation images.
The evaluation sets described below were generated in a separate batch
after the training pool had been frozen. To reduce exact and near-duplicate
leakage, we compare evaluation samples against the training pool using
exact-image hashes, perceptual-image similarity, and normalized
text/prompt matching.

\paragraph{Annotation protocol.}
Twenty annotators completed a three-round qualification procedure before
labeling data. Each image was independently annotated twice using a
customized CVAT interface, and disagreements were resolved through expert
arbitration. This protocol yields a Fleiss' $\kappa$ of $0.82$ for
categorical annotations and a mean pairwise spatial IoU of $0.79$ for
bounding boxes.

\paragraph{Grounded Reverse-CoT synthesis.}
Bounding boxes identify anomalous regions but do not directly provide the
reasoning format required by a diagnostic MLLM. We therefore construct a
\textbf{Reverse Chain-of-Thought (Reverse-CoT)} subset by supplying
Gemini 3.1 Pro with human-verified anomaly labels and bounding boxes and
asking it to explain the corresponding structural inconsistency. Unlike
open-ended artifact discovery, this procedure conditions the rationale on
verified spatial targets, reducing unsupported localization during data
synthesis.

The original text-to-image prompt is hidden from the teacher model so
that the explanation must be derived from visual evidence rather than
prompt-specific shortcuts. Absolute pixel boxes are converted to
normalized permille coordinates using the format
\texttt{<box>[[ymin, xmin, ymax, xmax]]</box>}, matching the coordinate
representation used during ARGUS training. We further remove examples
whose structural validity cannot be meaningfully assessed, including
purely fictional anatomy and severely obscured or underexposed subjects.
After automatic filtering and manual verification, 10K grounded
Reverse-CoT examples are selected from the training split for supervised
cold-start training.

\subsection{The Achilles-Benchmark}
\label{sec:achilles_benchmark}

The diagnostic component of Achilles-Benchmark contains 1,383
human-verified images, comprising 1,185 hard-positive examples and
198 normal controls, with 3,064 annotated anomaly boxes in total.
The benchmark emphasizes localized failures rather than global semantic
quality: its median bounding-box area is approximately
$1{,}224~\mathrm{px}^{2}$, and its long-tailed distribution includes
small human features, animal anatomy, object morphology, and
cross-entity interactions. All benchmark images are excluded from both
the supervised and reinforcement-learning stages.

For end-to-end generator evaluation, we construct 800
prompts balanced across humans, animals, interactions, and complex
objects. The prompts vary environments, viewpoints, and photographic
styles while explicitly requiring the structurally challenging subject
or body part to remain visible. A subject-verification step
checks whether the requested entity is present; outputs that evade the
structural requirement by omitting or hiding the target are counted as
prompt-alignment failures. For the remaining outputs, structural fidelity
is measured by the defect-free rate,
\begin{equation}
    S_{\mathrm{fidelity}}
    =
    100\left(1-\frac{N_{\mathrm{defect}}}{N_{\mathrm{total}}}\right),
\end{equation}
where $N_{\mathrm{defect}}$ is the number of verified images with any
structural anomaly and $N_{\mathrm{total}}$ is the number of evaluated
generations.
\section{The Diagnose-to-Generate Framework}
\label{sec:methods}

\begin{figure*}[t]
    \centering
    \includegraphics[width=\linewidth]{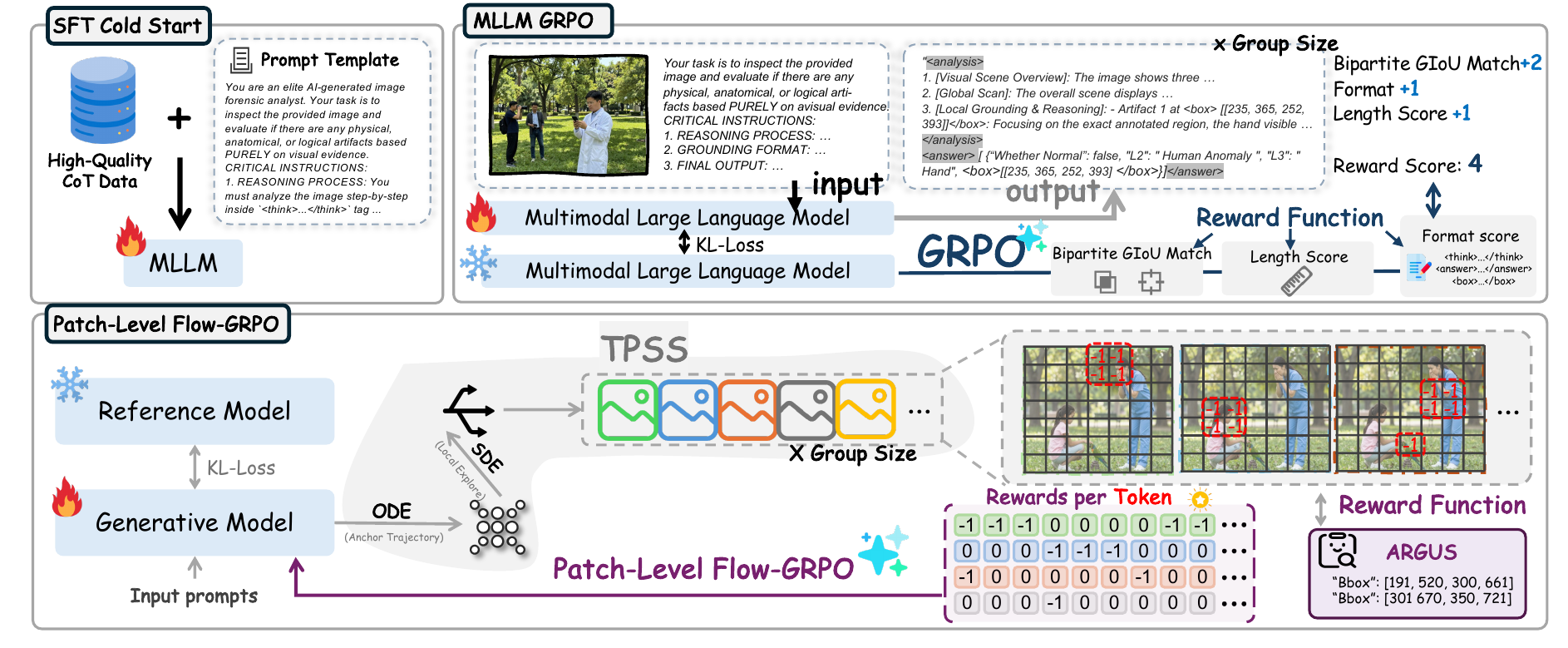}
    % \vspace{-4mm}
    \caption{\textbf{Overview of Diagnose-to-Generate (D2G).}
    Stage I trains ARGUS via grounded SFT and Region-Aware GRPO
    to localize structural anomalies. Stage II uses ARGUS diagnoses
    to spatially supervise Patch-level Flow-GRPO. TPSS preserves rollout
    correspondence, enabling localized advantages to update diagnosed regions
    while limiting changes elsewhere.}
    \label{fig:method_framework}
    % \vspace{-2mm}
\end{figure*}

D2G separates localized structural alignment into two stages, as
illustrated in Figure~\ref{fig:method_framework}. In Stage I, we train
\textbf{ARGUS}, a diagnostic specialist that predicts the category,
location, and explanation of structural anomalies. In Stage II, ARGUS
is frozen and used as a spatial reward emitter for
\textbf{Patch-level Flow-GRPO}. The latter preserves spatial
correspondence across rollout groups and converts predicted anomaly
regions into localized policy updates.

\subsection{Stage I: Grounded Diagnosis with ARGUS}
\label{sec:argus}

\paragraph{Grounded SFT initialization.}
Direct reinforcement learning is ineffective when the initial VLM rarely
produces valid fine-grained coordinates. We therefore initialize ARGUS
through supervised fine-tuning on the 10K Reverse-CoT examples in
Section~\ref{sec:data_construction}, following the cold-start strategy
of reasoning-oriented alignment~\cite{deepseekr1}. Each training target
contains a visual diagnosis, hierarchical anomaly labels, and normalized
bounding boxes. This stage teaches the diagnostic taxonomy, output format,
and correspondence between structural evidence and spatial coordinates.

\paragraph{Region-Aware GRPO.}
Starting from the SFT model, we apply GRPO to improve diagnostic
correctness and localization. For each image--instruction pair, the policy
samples a group of responses, whose rewards are normalized within the
group and optimized using the standard clipped GRPO objective with a KL
penalty to the reference policy~\cite{deepseekr1}. Our contribution lies
in the region-aware reward used to evaluate each response.

An image may contain multiple anomalies, and their textual output order
is not semantically meaningful. Let
$\hat{\mathcal{B}}=\{\hat{b}_j\}_{j=1}^{N_p}$ and
$\mathcal{B}=\{b_k\}_{k=1}^{N_g}$ denote the predicted and ground-truth
box sets. Inspired by set-based prediction~\cite{carion2020end}, we use
the Hungarian algorithm to obtain the assignment
\begin{equation}
\label{eq:hungarian}
    \hat{\sigma}
    =
    \arg\max_{\sigma}
    \sum_{k=1}^{K}
    \operatorname{GIoU}
    \bigl(\hat{b}_{\sigma(k)},b_k\bigr),
    \qquad
    K=\min(N_p,N_g),
\end{equation}
where GIoU provides informative feedback even when two boxes do not
overlap~\cite{rezatofighi2019generalized}.

The resulting sequence reward combines an output-format reward $r_0$,
the matched localization reward $r_{\mathrm{loc}}$, and hierarchical
classification rewards $(r_1,r_2,r_3)$. To prevent a response from
receiving classification credit without correctly grounding its evidence,
we introduce a localization consistency gate:
\begin{equation}
\label{eq:argus_reward}
    R
    =
    r_0+r_{\mathrm{loc}}
    +r_c(r_1+r_2+r_3),
    \qquad
    r_c
    =
    \mathbb{I}
    \left[
        \overline{\operatorname{IoU}}_{\hat{\sigma}}
        \geq \tau_{\mathrm{IoU}}
    \right].
\end{equation}
Thus, category rewards are enabled only when the matched boxes satisfy
the localization threshold. This design discourages reward strategies
that predict plausible anomaly labels without identifying the
corresponding visual evidence.

\paragraph{Distribution rebalancing.}
Structural anomalies follow a long-tailed distribution, while normal
images and frequent human-hand errors can dominate random batches.
We therefore use multi-bucket sampling to balance normal examples and
the four primary anomaly categories. We additionally upsample hard
normal examples containing visible yet anatomically valid structures.
This prevents ARGUS from learning shortcuts such as treating every
visible hand or limb as anomalous.

\subsection{Stage II: Patch-level Flow-GRPO}
\label{sec:patch_flow_grpo}

We use the frozen ARGUS model to align a Flow Matching
generator~\cite{flow_matching,liu2022flow}. Existing Flow-GRPO methods
introduce SDE sampling into deterministic flow trajectories to obtain
online rollouts~\cite{liu2025flow,dancegrpo,song2020score}.
Although effective for image-level rewards, applying stochasticity over
the entire trajectory can change object positions and shapes across
samples in the same group. A coordinate predicted in one rollout may
refer to a different semantic region in another, making patch-wise
relative advantages unreliable.

\paragraph{Trajectory-Preserving Stochastic Sampling.}
We address this problem with
\textbf{Trajectory-Preserving Stochastic Sampling (TPSS)}.
For each prompt, the rollout group shares an early deterministic ODE
trajectory that establishes the scene layout. The trajectory branches
into $G$ stochastic rollouts only after a critical timestep
$t_{\mathrm{critical}}\in(0,1)$. For rollout $i$, the reverse-time update
is
\begin{equation}
\label{eq:tpss}
    x^{(i)}_{t-\Delta t}
    =
    x^{(i)}_t
    -
    v_{\theta}\!\left(x^{(i)}_t,t\right)\Delta t
    +
    \mathbb{I}
    \left[t<t_{\mathrm{critical}}\right]
    \sigma_t\sqrt{\Delta t}\,
    \epsilon^{(i)}_t,
\end{equation}
where $\epsilon^{(i)}_t\sim\mathcal{N}(0,I)$. The early interval
$t\in[t_{\mathrm{critical}},1]$ is therefore deterministic, while
stochastic exploration is restricted to the late interval
$t<t_{\mathrm{critical}}$.

TPSS does not preserve exactly the same sampling marginal as a full SDE
trajectory. Instead, it deliberately shortens the stochastic interval to
trade unrestricted exploration for stronger spatial correspondence.
Because the macro-layout is established before the trajectories branch,
late perturbations primarily affect local appearance and morphology.
We quantify the resulting residual drift and the sensitivity to
$t_{\mathrm{critical}}$ in Section~\ref{sec:experiments}.

\paragraph{From grounded boxes to dense spatial rewards.}
After decoding rollout $i$, ARGUS predicts a set of anomaly boxes
$\mathcal{B}_i$ and a diagnostic confidence $c_i$. The boxes are mapped
to the latent-token grid and rasterized into a spatial mask. To avoid
sharp optimization boundaries, we Gaussian-smooth the box support to
obtain
$M_i^{(h,w)}\in[0,1]$, where larger values indicate tokens closer to
the diagnosed region.

ARGUS may produce unreliable boxes in heavily occluded or ambiguous
images. We therefore apply a confidence safeguard:
\begin{equation}
\label{eq:confidence_gate}
    \lambda_i
    =
    \lambda_p\,
    \mathbb{I}[c_i\geq\tau_c],
    \qquad \tau_c=0.3,
\end{equation}
so low-confidence diagnoses contribute no structural penalty and the
optimization falls back to the base reward and reference-policy
regularization. The dense reward at latent position $(h,w)$ is
\begin{equation}
\label{eq:dense_reward}
    R_i^{(h,w)}
    =
    \mathcal{R}^{(i)}_{\mathrm{base}}
    -
    \lambda_i M_i^{(h,w)},
\end{equation}
where $\mathcal{R}^{(i)}_{\mathrm{base}}$ is an image-level quality reward
provided by an off-the-shelf scorer such as
ImageReward~\cite{xu2023imagereward}. For every latent position, we
normalize rewards across the $G$ rollouts:
\begin{equation}
\label{eq:dense_advantage}
    \hat{A}_i^{(h,w)}
    =
    \frac{
        R_i^{(h,w)}-\mu_R^{(h,w)}
    }{
        \sigma_R^{(h,w)}+\epsilon
    },
\end{equation}
where $\mu_R^{(h,w)}$ and $\sigma_R^{(h,w)}$ are the group-wise mean
and standard deviation.

\paragraph{Localized gradient modulation.}
Following the flow-policy gradient formulation of
Flow-GRPO~\cite{liu2025flow}, we use the dense advantage to modulate
the generator update:
\begin{equation}
\label{eq:patch_gradient}
    \nabla_{\theta}\mathcal{L}_{\mathrm{patch}}
    =
    -w_t
    \sum_{h,w}
    M_i^{(h,w)}
    \hat{A}_i^{(h,w)}
    \epsilon_i^{(h,w)}
    \nabla_{\theta}
    v_{\theta}^{(h,w)} .
\end{equation}
The explicit factor $M_i^{(h,w)}$ makes the patch branch contribute
zero gradient outside the diagnosed support, while Gaussian smoothing
provides a gradual transition near box boundaries. We retain the
reference-model KL regularization of Flow-GRPO to constrain deviation
from the pretrained generator. Following the score-gradient scaling of
the adopted SDE kernel~\cite{liu2025flow}, we use the stable coefficient
$w_t=3/2$ throughout training.
\section{Experiments}
\label{sec:experiments}

We evaluate D2G from three perspectives: 
(1) whether ARGUS accurately detects and localizes structural anomalies;
(2) whether Patch-level Flow-GRPO improves structural fidelity without
degrading global image quality; and 
(3) whether the proposed data, reward, and sampling designs account for
the observed improvements.

\subsection{Experimental Setup}
\label{sec:exp_setup}

\paragraph{Models and optimization.}
For Stage I, ARGUS is initialized from Qwen3-VL-8B~\cite{qwen3vl}.
We train variants based on Qwen2.5-VL-7B
~\cite{bai2025qwen25vltechnicalreport} and InternVL3.5-8B
~\cite{wang2025internvl3} to assess backbone generalization.
The SFT cold-start uses a batch size of 128 for two epochs,
with a cosine schedule peaking at $2\times10^{-5}$. Region-Aware
GRPO uses a group size of $G=8$, a learning rate of
$1\times10^{-6}$, a KL coefficient of $\beta=0.01$, and an IoU
consistency threshold of $0.5$.

For Stage II, we optimize the transformer backbone of
FLUX.1-dev~\cite{flux2024} for 1,000 iterations. We use $G=4$
rollouts per prompt, a learning rate of $5\times10^{-6}$,
$t_{\mathrm{critical}}=0.3$, and a Gaussian mask radius of
$\sigma=1.5$ latent pixels. Experiments are implemented with
PyTorch and DeepSpeed ZeRO-2 on 32 NVIDIA H800 GPUs. Stage I
requires approximately 68 hours, including 4 hours for SFT and
64 hours for GRPO, while one Stage-II run requires approximately
18 hours. We repeat Stage II with three random seeds.

\paragraph{Evaluation protocols.}
ARGUS is evaluated on the 1,383-image diagnostic split of
Achilles-Benchmark. We report binary anomaly detection, L2 Macro
F1, spatial mIoU, and human-alignment accuracy. L2 Macro F1 is the
mean over four top-level anomaly categories.
For comparable spatial evaluation, GPT-4.1~\cite{GPT-4.1}
and Gemini 3.1 Pro~\cite{Gemini-3-pro} are prompted with structured
examples and required to output
\texttt{[ymin,xmin,ymax,xmax]} coordinates.

Generative structural fidelity is measured on the 800-prompt
Achilles-Benchmark. Global preservation is evaluated using FID-30K
~\cite{FID} and CLIP Score~\cite{hessel2021clipscore} on 30,000
MS-COCO prompts. To reduce dependence on ARGUS, we also use
HumanRefiner~\cite{HumanRefiner}, zero-shot GPT-4.1, and a
double-blind human study. The study contains 300 randomly sampled
pairs evaluated by 15 experts, with three raters per pair and
majority voting; inter-rater agreement is Fleiss' $\kappa=0.78$.
Unless otherwise specified, bold and underlined values denote the
best and second-best results.

\begin{figure*}[t]
    \centering
    \includegraphics[width=0.98\linewidth]
    {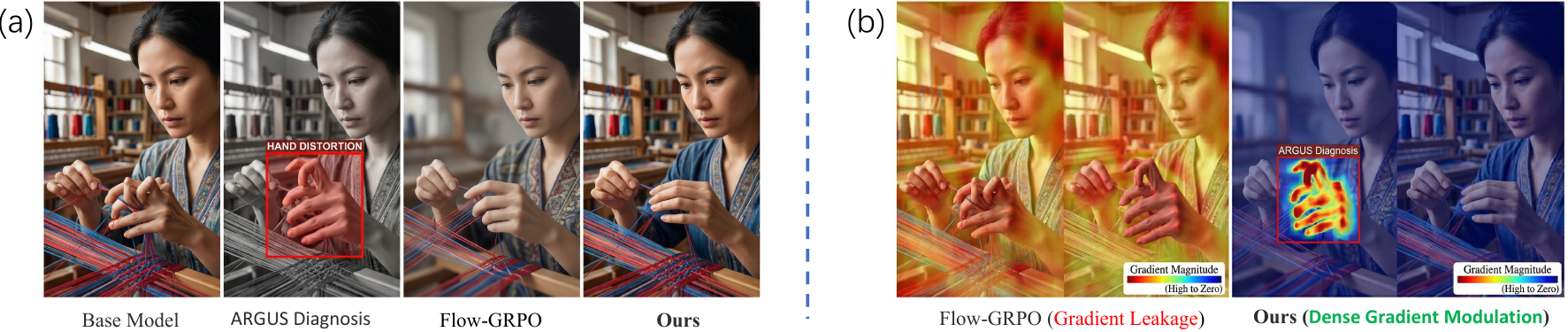}
    % \vspace{-3mm}
    \caption{\textbf{Qualitative comparison of generative alignment.}
    \textbf{(a)} Global Flow-GRPO alters surrounding appearance, whereas D2G preserves it during local correction.
    \textbf{(b)} Update magnitudes are globally distributed for Flow-GRPO but concentrated around ARGUS-diagnosed regions for D2G.}
    \label{fig:qualitative_visuals}
    % \vspace{-2mm}
\end{figure*}

\subsection{Grounded Diagnostic Performance}
\label{sec:argus_results}

\begin{table*}[t]
\caption{
\textbf{Diagnostic performance on Achilles-Benchmark.}
L2 Macro averages precision, recall, and F1 over four anomaly categories;
Binary evaluates anomaly detection, and mIoU evaluates localization.
$\dagger$ denotes structured CoT prompting for box outputs.
Human Align. follows~\cite{wang2026thinking}.
}
\label{table:main result}
% \vspace{-2mm}
\centering
\resizebox{0.86\textwidth}{!}{
\renewcommand{\arraystretch}{1.12}
\begin{tabular}{ll ccc ccc cc}
\toprule
\multirow{2}{*}{\textbf{Category}}
& \multirow{2}{*}{\textbf{Model}}
& \multicolumn{3}{c}{\textbf{L2 Labels Macro}}
& \multicolumn{3}{c}{\textbf{Artifact Detection}}
& \textbf{Spatial}
& \textbf{Human Align.} \\
\cmidrule(lr){3-5}
\cmidrule(lr){6-8}
& & \textbf{Prec.} & \textbf{Rec.} & \textbf{F1}
& \textbf{Prec.} & \textbf{Rec.} & \textbf{F1}
& \textbf{mIoU ($\uparrow$)}
& \textbf{Acc. (\%) ($\uparrow$)} \\
\midrule

\multirow{2}{*}{\textit{Proprietary APIs}}
& GPT-4.1$\dagger$~\cite{GPT-4.1}
& 0.4452 & 0.2105 & 0.2855
& 0.6315 & 0.4055 & 0.4950
& 0.2361 & 68.5 \\

& Gemini 3.1 Pro$\dagger$~\cite{Gemini-3-pro}
& 0.4294 & 0.3572 & 0.3905
& 0.5577 & \textbf{0.8214} & 0.6631
& 0.2814 & 72.4 \\
\midrule

\multirow{3}{*}{\textit{Zero-Shot Foundation}}
& Qwen2.5-VL-7B~\cite{bai2025qwen25vltechnicalreport}
& 0.3395 & 0.0995 & 0.1538
& 0.5674 & 0.3597 & 0.4403
& -- & 55.2 \\

& InternVL3.5-8B~\cite{wang2025internvl3}
& 0.3802 & 0.1205 & 0.1830
& 0.5810 & 0.3822 & 0.4611
& -- & 58.1 \\

& Qwen3-VL-8B~\cite{qwen3vl}
& 0.4105 & 0.1620 & 0.2325
& 0.6015 & 0.4105 & 0.4880
& -- & 61.3 \\
\midrule

\multirow{2}{*}{\textit{SOTA Evaluators}}
& MagicAssessor~\cite{wang2025magicmirror}
  \scriptsize{(Qwen2.5-7B)}
& 0.5446 & 0.5244 & 0.5261
& 0.7756 & 0.6381 & 0.7001
& -- & 75.8 \\

& REACT~\cite{wang2026thinking}
  \scriptsize{(Qwen2.5-7B)}
& 0.5812 & 0.5530 & 0.5667
& 0.7920 & 0.6815 & 0.7325
& -- & 79.2 \\
\midrule

\multirow{3}{*}{\textit{Ours}}
& \textbf{ARGUS} \scriptsize{(Qwen2.5-VL-7B)}
& 0.6420 & 0.6215 & 0.6315
& 0.8350 & 0.7510 & 0.7907
& 0.6845 & 82.4 \\

& \textbf{ARGUS} \scriptsize{(InternVL3.5-8B)}
& \underline{0.6515}
& \underline{0.6308}
& \underline{0.6409}
& \underline{0.8412}
& 0.7605
& \underline{0.7988}
& \underline{0.6912}
& \underline{83.1} \\

& \textbf{ARGUS} \scriptsize{(Qwen3-VL-8B)}
& \textbf{0.6845}
& \textbf{0.6630}
& \textbf{0.6735}
& \textbf{0.8655}
& \underline{0.8012}
& \textbf{0.8321}
& \textbf{0.7540}
& \textbf{86.7} \\
\bottomrule
\end{tabular}
}
% \vspace{-2mm}
\end{table*}

Table~\ref{table:main result} compares ARGUS with proprietary APIs,
zero-shot VLMs, and dedicated image evaluators. ARGUS based on
Qwen3-VL-8B achieves a binary F1 of $0.8321$, an L2 Macro F1 of
$0.6735$, an mIoU of $0.7540$, and an $86.7\%$ human-alignment
accuracy. The improvements are consistent across backbones:
the Qwen2.5-VL and InternVL3.5 variants obtain binary F1 scores of
$0.7907$ and $0.7988$, respectively, with mIoU above $0.68$.

Requiring proprietary models to emit boxes does not close the
grounding gap. GPT-4.1 obtains precision/recall of
$0.6315/0.4055$ but only $0.2361$ mIoU, while Gemini 3.1 Pro
obtains $0.5577/0.8214$ and $0.2814$ mIoU. In contrast, ARGUS
reaches $0.8655/0.8012$ precision/recall and $0.7540$ mIoU.
Thus, the benefit is not explained by coordinate prompting alone;
dedicated region-aware training is important for fine-grained
localization. MagicAssessor~\cite{wang2025magicmirror} and
REACT~\cite{wang2026thinking} provide stronger global judgments
than zero-shot VLMs, but do not provide comparably precise
box-level supervision.

\begin{table*}[t]
\caption{
\textbf{Diagnostic performance across fine-grained distortion categories.}
We report precision, recall, and F1 for abnormal interaction, human anatomy,
animal anatomy, and object morphology. ARGUS consistently improves
category-level F1 across three underlying VLM backbones.
}
\label{table:main result 4 class}
% \vspace{-2mm}
\centering
\resizebox{0.86\textwidth}{!}{
\renewcommand{\arraystretch}{1.10}
\begin{tabular}{ll ccc ccc ccc ccc}
\toprule
\multirow{2}{*}{\textbf{Category}}
& \multirow{2}{*}{\textbf{Model}}
& \multicolumn{3}{c}{\textbf{Abnormal Interaction}}
& \multicolumn{3}{c}{\textbf{Human Anatomy}}
& \multicolumn{3}{c}{\textbf{Animal Anatomy}}
& \multicolumn{3}{c}{\textbf{Object Morphology}} \\
\cmidrule(lr){3-5}
\cmidrule(lr){6-8}
\cmidrule(lr){9-11}
\cmidrule(lr){12-14}
& & \textbf{Prec.} & \textbf{Rec.} & \textbf{F1}
& \textbf{Prec.} & \textbf{Rec.} & \textbf{F1}
& \textbf{Prec.} & \textbf{Rec.} & \textbf{F1}
& \textbf{Prec.} & \textbf{Rec.} & \textbf{F1} \\
\midrule

\multirow{2}{*}{\textit{Proprietary APIs}}
& GPT-4.1~\cite{GPT-4.1}
& 0.3112 & 0.2054 & 0.2478
& 0.7305 & 0.3415 & 0.4655
& 0.5840 & 0.2512 & 0.3512
& 0.2645 & 0.1130 & 0.1583 \\

& Gemini 3.1 Pro~\cite{Gemini-3-pro}
& 0.3204 & 0.3015 & 0.3106
& 0.7645 & 0.5012 & 0.6054
& 0.5620 & 0.3840 & 0.4562
& 0.2215 & 0.4420 & 0.2885 \\
\midrule

\multirow{3}{*}{\textit{Zero-Shot Found.}}
& Qwen2.5-VL-7B~\cite{bai2025qwen25vltechnicalreport}
& 0.2008 & 0.1180 & 0.1486
& 0.6182 & 0.1358 & 0.2227
& 0.4361 & 0.1005 & 0.1634
& 0.1028 & 0.0436 & 0.0612 \\

& InternVL3.5-8B~\cite{wang2025internvl3}
& 0.2315 & 0.1350 & 0.1705
& 0.6410 & 0.1620 & 0.2586
& 0.4720 & 0.1245 & 0.1965
& 0.1345 & 0.0620 & 0.0848 \\

& Qwen3-VL-8B~\cite{qwen3vl}
& 0.2550 & 0.1640 & 0.1988
& 0.6855 & 0.2105 & 0.3222
& 0.5105 & 0.1660 & 0.2505
& 0.1802 & 0.0915 & 0.1215 \\
\midrule

\multirow{2}{*}{\textit{SOTA Evaluators}}
& MagicAssessor~\cite{wang2025magicmirror}
& 0.3665 & 0.4621 & 0.4088
& 0.8957 & 0.6177 & 0.7312
& 0.5805 & 0.6563 & 0.6161
& 0.3359 & 0.3616 & 0.3482 \\

& REACT~\cite{wang2026thinking}
& 0.4215 & 0.5012 & 0.4578
& \underline{0.9105} & 0.6540 & 0.7615
& 0.6210 & 0.6955 & 0.6542
& 0.3850 & 0.4125 & 0.3982 \\
\midrule

\multirow{3}{*}{\textit{Ours}}
& \textbf{ARGUS} \scriptsize{(Qwen2.5-VL)}
& 0.4855 & 0.5620 & 0.5209
& 0.9025 & 0.7850 & 0.8397
& 0.6845 & 0.7510 & 0.7162
& 0.4625 & 0.4855 & 0.4737 \\

& \textbf{ARGUS} \scriptsize{(InternVL3.5)}
& \underline{0.5012}
& \underline{0.5745}
& \underline{0.5353}
& 0.9088
& \underline{0.8015}
& \underline{0.8518}
& \underline{0.6950}
& \underline{0.7625}
& \underline{0.7271}
& \underline{0.4710}
& \underline{0.5020}
& \underline{0.4860} \\

& \textbf{ARGUS} \scriptsize{(Qwen3-VL)}
& \textbf{0.5420}
& \textbf{0.6210}
& \textbf{0.5788}
& \textbf{0.9245}
& \textbf{0.8520}
& \textbf{0.8867}
& \textbf{0.7315}
& \textbf{0.8050}
& \textbf{0.7664}
& \textbf{0.5215}
& \textbf{0.5540}
& \textbf{0.5372} \\
\bottomrule
\end{tabular}
}
% \vspace{-2mm}
\end{table*}

Table~\ref{table:main result 4 class} further reports all
fine-grained categories. ARGUS with Qwen3-VL obtains F1 scores of
$0.5788$, $0.8867$, $0.7664$, and $0.5372$ for abnormal
interaction, human anatomy, animal anatomy, and object morphology,
respectively. The corresponding REACT results are $0.4578$,
$0.7615$, $0.6542$, and $0.3982$. The consistent gains across both
frequent human anomalies and less represented interaction, animal,
and object categories indicate that the improvement is not confined
to hand-specific detection.

\subsection{End-to-End Generative Alignment}
\label{sec:generative_results}

\begin{table*}[t]
\caption{
\textbf{End-to-end structural alignment and independent evaluation.}
Panels A--B report Achilles-Benchmark structural fidelity and global
preservation. Panel C evaluates FLUX.1-dev outputs using HumanRefiner,
zero-shot GPT-4.1, and double-blind human preference, and includes local
alignment baselines. Panel D evaluates cross-architecture transfer on
SD3.5-M. Best and second-best results are marked within each panel.
}
\label{tab:generative_fidelity}
% \vspace{-2mm}
\centering

% ------------------------------------------------------------------
% Panels A and B
% ------------------------------------------------------------------
\resizebox{0.86\textwidth}{!}{
\renewcommand{\arraystretch}{1.08}
\begin{tabular}{l ccccc ccc}
\toprule
\multirow{2}{*}{\textbf{Model / Alignment Strategy}}
& \multicolumn{5}{c}{\textbf{Structural Fidelity on Achilles-Benchmark ($\uparrow$)}}
& \multicolumn{3}{c}{\textbf{Global Preservation and Human Alignment}} \\
\cmidrule(lr){2-6}
\cmidrule(lr){7-9}
& \textbf{Interact.}
& \textbf{Human}
& \textbf{Animal}
& \textbf{Object}
& \textbf{Overall}
& \textbf{FID-30K ($\downarrow$)}
& \textbf{CLIP ($\uparrow$)}
& \textbf{Human Win ($\uparrow$)} \\
\midrule

\multicolumn{9}{l}{
\textit{Panel A: Contemporary Foundation Models (Zero-Shot)}
} \\

GPT-image-1~\cite{chen2025opengpt4oimagecomprehensivedatasetadvanced}
& 81.54
& \textbf{45.00}
& \textbf{49.00}
& \textbf{91.41}
& \textbf{63.08}
& -- & --
& \underline{52.4\%} \\

HunyuanVideo~\cite{kong2024hunyuanvideo}
& \textbf{86.15}
& \underline{43.20}
& \underline{48.50}
& 88.50
& \underline{62.15}
& -- & --
& 51.0\% \\

Bagel~\cite{deng2505emerging}
& \underline{85.96}
& 41.00
& 48.00
& 87.92
& 60.53
& -- & --
& \textbf{53.1\%} \\

Seedance 1.0 Pro~\cite{gao2025seedance}
& 78.37
& 41.00
& 43.00
& \underline{90.56}
& 59.54
& -- & --
& 48.5\% \\
\midrule

\multicolumn{9}{l}{
\textit{Panel B: Stage-II Alignment on FLUX.1-dev}
} \\

Base Model (FLUX.1-dev)
& 84.71
& 46.00
& 44.50
& 89.60
& 62.16
& \underline{18.42}
& \underline{32.15}
& 50.0\% \\

\quad + SFT (High-Quality Subset)
& 86.20
& 51.40
& 49.30
& 90.15
& 66.85
& 19.35
& 31.80
& \underline{55.4\%} \\

\quad + Flow-GRPO~\cite{liu2025flow}
& 87.15
& 55.30
& 54.10
& 90.45
& 69.45
& 24.50
& 29.80
& 42.1\% \\

\quad + DanceGRPO~\cite{dancegrpo}
& \underline{88.05}
& \underline{58.65}
& \underline{57.20}
& \underline{91.30}
& \underline{71.60}
& 22.10
& 30.55
& 47.8\% \\

\quad \textbf{+ Patch-level Flow-GRPO (Ours)}
& \textbf{94.10}
& \textbf{81.25}
& \textbf{78.60}
& \textbf{96.55}
& \textbf{85.60}
& \textbf{18.35}
& \textbf{32.22}
& \textbf{82.5\%} \\
\bottomrule
\end{tabular}
}

\vspace{1.5mm}

\hspace{-9mm}
% ------------------------------------------------------------------
% Panels C and D
% ------------------------------------------------------------------
\begin{minipage}[t]{0.61\textwidth}
\centering
\resizebox{0.76\linewidth}{!}{
\renewcommand{\arraystretch}{1.08}
\begin{tabular}{l cccc}
\toprule
\multicolumn{5}{l}{
\textit{Panel C: Independent Evaluation and Local Baselines on FLUX.1-dev}
} \\
\textbf{Alignment Strategy}
& \shortstack{\textbf{HumanRefiner}\\\textbf{AFR ($\uparrow$)}}
& \shortstack{\textbf{GPT-4.1}\\\textbf{Win ($\uparrow$)}}
& \shortstack{\textbf{Human}\\\textbf{Win ($\uparrow$)}}
& \shortstack{\textbf{FID-30K}\\\textbf{($\downarrow$)}} \\
\midrule

Base Model
& 42.2\%
& 50.0\%
& 50.0\%
& \underline{18.42} \\

+ Global Flow-GRPO~\cite{liu2025flow}
& 44.5\%
& 43.6\%
& 42.1\%
& 24.50 \\

+ Masked Inpainting
& \underline{68.2\%}
& \underline{64.3\%}
& \underline{62.5\%}
& 21.64 \\

+ adapted-LocalDPO~\cite{localdpo}
& 60.5\%
& 56.8\%
& 58.3\%
& 20.27 \\

\textbf{+ Patch-level Flow-GRPO (Ours)}
& \textbf{74.8\%}
& \textbf{80.2\%}
& \textbf{82.5\%}
& \textbf{18.35} \\
\bottomrule
\end{tabular}
}
\end{minipage}
% \hfill
\begin{minipage}[t]{0.37\textwidth}
\centering
\resizebox{\linewidth}{!}{
\renewcommand{\arraystretch}{1.08}
\begin{tabular}{l ccc}
\toprule
\multicolumn{4}{l}{
\textit{Panel D: Cross-Architecture Results on SD3.5-M}
} \\
\textbf{Alignment Strategy}
& \shortstack{\textbf{Human}\\\textbf{Fidelity}}
& \shortstack{\textbf{Overall}\\\textbf{Fidelity}}
& \shortstack{\textbf{FID-30K}\\\textbf{($\downarrow$)}} \\
\midrule

Base Model~\cite{sd3}
& 39.33
& 54.82
& \underline{20.15} \\

+ Global Flow-GRPO~\cite{liu2025flow}
& \underline{42.10}
& \underline{58.20}
& 26.40 \\

\textbf{+ Patch-level Flow-GRPO}
& \textbf{76.45}
& \textbf{81.50}
& \textbf{20.10} \\
\bottomrule
\end{tabular}
}
\end{minipage}

% \vspace{-2mm}
\end{table*}

\paragraph{Structural correction and global preservation.}
Table~\ref{tab:generative_fidelity} reports end-to-end results.
The zero-shot models in Panel A obtain structural fidelity
between $59.54$ and $63.08$, showing that localized geometric
failures remain common despite strong global synthesis quality.

On FLUX.1-dev, SFT improves overall fidelity from $62.16$ to
$66.85$, but slightly degrades FID and CLIP Score. Global
Flow-GRPO~\cite{liu2025flow} and DanceGRPO~\cite{dancegrpo}
increase fidelity to $69.45$ and $71.60$, respectively, while
degrading FID from $18.42$ to $24.50$ and $22.10$. Their human
preference rates are also below the base-model reference
($42.1\%$ and $47.8\%$).

Patch-level Flow-GRPO reaches $85.60$ overall fidelity, including
$94.10$, $81.25$, $78.60$, and $96.55$ on interaction, human,
animal, and object categories. Meanwhile, FID remains $18.35$
versus $18.42$ for the base model, and CLIP Score increases from
$32.15$ to $32.22$. These results indicate that the structural gain
does not require a measurable loss of global distribution quality or
prompt correspondence.

\paragraph{Independent evaluation and local baselines.}
Panel C evaluates the same FLUX.1-dev variants without using ARGUS
as the judge. Under HumanRefiner, the anomaly-free rates of the
base model and Global Flow-GRPO are $42.2\%$ and $44.5\%$.
Masked inpainting improves this metric to $68.2\%$, but degrades
FID to $21.64$. We also adapt the masked preference mechanism of
LocalDPO~\cite{localdpo} by replacing its random masks with ARGUS
semantic masks, providing it with the same detector-derived spatial
prior used by D2G. This strengthened baseline obtains a $60.5\%$
anomaly-free rate, a $56.8\%$ GPT-4.1 win rate, a $58.3\%$ human
win rate, and an FID of $20.27$.

D2G obtains the strongest independent results: $74.8\%$
HumanRefiner anomaly-free rate, $80.2\%$ GPT-4.1 win rate, and
$82.5\%$ human preference, while retaining an FID of $18.35$.
HumanRefiner uses a stricter human-centric taxonomy than ARGUS,
so its absolute score is lower than the $85.60$ Achilles fidelity;
nevertheless, both evaluators show a similar relative improvement
over the base model.

\paragraph{OOD and cross-architecture generalization.}
On 1,000 unseen human-centric MS-COCO prompts, HumanRefiner
reports a $71.5\%$ anomaly-free rate for D2G, compared with
$41.8\%$ for the base model. This set is distinct from the
Achilles evaluation prompts and therefore tests transfer beyond the
training and benchmark distributions.

We also apply the complete Stage-II procedure to Stable Diffusion
3.5-Medium~\cite{sd3}. Its base model obtains $39.33$ human
fidelity, $54.82$ overall fidelity, and an FID of $20.15$.
Global Flow-GRPO improves overall fidelity only to $58.20$ while
degrading FID to $26.40$. Patch-level Flow-GRPO instead reaches
$76.45$ human fidelity and $81.50$ overall fidelity, with an FID of
$20.10$. The similar behavior on FLUX.1-dev and SD3.5-M suggests
that the localized optimization is not tied to one generator
architecture.

\subsection{Mechanistic Analysis}
\label{sec:mechanistic_results}

\begin{table}[t]
\caption{
\textbf{Component ablations.}
Stage I reports diagnostic F1 and localization mIoU; Stage II reports
Achilles structural fidelity and FID-30K. Reverse-CoT evaluates the
grounded data engine, while Confidence-Decay evaluates robustness to
uncertain ARGUS diagnoses.
}
\label{tab:ablation}
% \vspace{-2mm}
\centering
\resizebox{\columnwidth}{!}{
\renewcommand{\arraystretch}{1.12}
\begin{tabular}{l cccc}
\toprule
\textbf{Ablation Variant}
& \textbf{Diag. F1}
& \textbf{mIoU}
& \textbf{Fidelity}
& \textbf{FID-30K} \\
\midrule

\multicolumn{5}{l}{
\textit{Stage I: Diagnostic Specialist}
} \\

\textbf{Full ARGUS}
& \textbf{0.8321}
& \textbf{0.7540}
& -- & -- \\

\quad w/o Reverse-CoT
& 0.7510
& 0.6150
& -- & -- \\

\quad w/o Bipartite GIoU Match
& 0.6845
& 0.5210
& -- & -- \\

\quad w/o Consistency Gate ($r_c$)
& 0.7890
& 0.4250
& -- & -- \\

\quad w/o Multi-Bucket Sampling
& 0.7415
& 0.7105
& -- & -- \\
\midrule

\multicolumn{5}{l}{
\textit{Stage II: Generative Alignment}
} \\

\textbf{Full D2G}
& -- & --
& \textbf{85.60}
& \textbf{18.35} \\

\quad w/o TPSS ($t_{\mathrm{critical}}=1$)
& -- & --
& 70.25
& 22.45 \\

\quad w/o Gaussian Soft Mask
& -- & --
& 83.90
& 19.75 \\

\quad w/o Confidence-Decay
& -- & --
& 84.15
& 21.40 \\
\bottomrule
\end{tabular}
}
% \vspace{-1mm}
\end{table}

\paragraph{Stage-I components.}
Table~\ref{tab:ablation} separates the effects of data synthesis,
matching, gating, and sampling. Replacing Reverse-CoT with
open-ended VQA reduces diagnostic F1 from $0.832$ to $0.751$
and mIoU from $0.754$ to $0.615$, indicating that verified spatial
priors reduce localization errors during cold-start training.
Removing Hungarian Bipartite GIoU matching further lowers F1 and
mIoU to $0.6845$ and $0.5210$. Without the consistency gate, the
model retains a deceptively high diagnostic F1 of $0.7890$, while
mIoU falls to $0.4250$, showing that category correctness alone does
not guarantee grounded evidence. Removing multi-bucket sampling
reduces F1 to $0.7415$, confirming the importance of balancing
normal and long-tailed anomaly categories.

\paragraph{TPSS and localized optimization.}
Without TPSS, early stochasticity disrupts correspondence among
group rollouts, reducing structural fidelity from $85.60$ to $70.25$
and degrading FID from $18.35$ to $22.45$. At
$t_{\mathrm{critical}}=0.3$, the mean displacement of matched
patches is $1.8\pm0.6$ pixels on a $1024^2$ canvas, compared with
more than 15 pixels under global SDE sampling. Performance remains
stable for $t_{\mathrm{critical}}\in[0.2,0.4]$; smaller thresholds
provide insufficient rollout diversity, whereas thresholds above
$0.6$ increase spatial drift and FID.

Replacing the Gaussian soft mask with a hard mask decreases
fidelity to $83.90$ and increases FID to $19.75$, consistent with
boundary artifacts introduced by abrupt latent transitions. Finally,
removing Confidence-Decay retains $84.15$ fidelity but degrades
FID to $21.40$, showing that suppressing low-confidence structural
penalties is important for preventing inaccurate ARGUS boxes from
altering otherwise valid regions.

\paragraph{Qualitative analysis.}
Figure~\ref{fig:qualitative_visuals} is consistent with the quantitative
results. Global Flow-GRPO can improve the malformed region but
also changes facial appearance, lighting, or background texture.
The ARGUS mask and patch-wise advantage concentrate the dominant
updates around the diagnosed region, producing a more localized
correction while preserving surrounding content.
\section{Conclusion}
\label{sec:conclusion}

We presented \textbf{Diagnose-to-Generate (D2G)}, a grounded alignment framework for correcting localized structural distortions in flow-based generators. D2G combines Achilles-600k and ARGUS for category-aware anomaly localization with Patch-level Flow-GRPO for spatially targeted optimization. Its Trajectory-Preserving Stochastic Sampling (TPSS) maintains correspondence across rollout groups by retaining deterministic early trajectories and restricting stochastic exploration to later denoising stages. Experiments across diagnostic backbones, generators, independent evaluators, and human studies show consistent improvements in structural fidelity while preserving global image quality. Future work will extend grounded alignment to video generation and further characterize the distributional effects of truncated stochastic sampling.

\bibliographystyle{ACM-Reference-Format}
\balance
\bibliography{mm26}

% \appendix
\clearpage
\appendix

\setcounter{table}{0}
\renewcommand{\thetable}{S\arabic{table}}
\setcounter{figure}{0}
\renewcommand{\thefigure}{S\arabic{figure}}
\setcounter{equation}{0}
\renewcommand{\theequation}{S\arabic{equation}}

% Avoid duplicate hyperlink destinations after resetting counters.
\providecommand{\theHtable}{}
\renewcommand{\theHtable}{supp.\arabic{table}}
\providecommand{\theHfigure}{}
\renewcommand{\theHfigure}{supp.\arabic{figure}}
\providecommand{\theHequation}{}
\renewcommand{\theHequation}{supp.\arabic{equation}}

\section{Dataset Construction and Annotation}
\label{supp:dataset}

\subsection{Image Sources and Annotation Taxonomy}

Achilles-600k contains images generated by Z-image~\cite{zimage}, Qwen-Image~\cite{qwen_image}, LongCat-Image~\cite{LongCat-Image}, and Nano Banana Pro~\cite{nano-banana-pro}. Generation prompts are synthesized using GPT-4.1~\cite{GPT-4.1} and Gemini 3.1 Pro~\cite{Gemini-3-pro}, with an emphasis on articulated subjects, object interactions, and compositions that expose structural errors. Using multiple generators broadens the observed artifact patterns, but does not by itself establish generalization to unseen generators.

The initial curation process uses six categories: \emph{normal images}, \emph{unreasonable element attributes}, \emph{unreasonable element interactions}, \emph{unreasonable physical structures}, \emph{other anomalies}, and \emph{discarded cases}. These collection-level labels should be distinguished from the four categories used to report diagnostic performance in the main paper: abnormal interaction, human anatomy, animal anatomy, and object morphology. The former organize image screening; the latter organize the evaluated structural defects. Localized anomalies receive individual bounding boxes rather than a single box enclosing all questionable content.

% AUTHOR CHECK: Include the actual mapping from collection labels to evaluation labels in the annotation release. The six-way curation taxonomy and four-way evaluation taxonomy must not be presented as the same label space.

\subsection{Data Splits and Annotation Quality}

The retained dataset contains 585K training images and 15K validation images. The 10K Reverse-CoT examples used for SFT are a subset of the training split, not an additional collection. The 1,383 diagnostic benchmark images and 800 generative evaluation prompts were curated in a separate generation batch after the training pool had been frozen. Evaluation samples were excluded from both training stages and checked against the training pool using exact-image hashes, perceptual-image deduplication, and normalized text/prompt matching. These checks reduce exact and near-duplicate leakage; they do not imply that the evaluation distribution is unrelated to the training distribution.

Twenty annotators completed a three-round qualification procedure. Every retained image was annotated twice using a customized CVAT interface, with disagreements resolved by expert arbitration. Agreement before arbitration was reported as Fleiss' $\kappa=0.82$ for categorical annotations and spatial mIoU of $0.79$ for bounding boxes. These annotation-agreement measurements are distinct from the diagnostic accuracy and localization mIoU reported for ARGUS.

% AUTHOR CHECK: Confirm whether agreement was measured before arbitration and specify its sampling scope. Record the deduplication thresholds, matching procedure, and split identifiers; these details are not recoverable from the aggregate statistics alone.

\subsection{Grounded Reverse-CoT Construction}

Reverse-CoT synthesis starts from human-verified anomaly labels and bounding boxes. Gemini 3.1 Pro is asked to explain the structural inconsistency at those locations rather than discover defects without spatial guidance. The original text-to-image prompt is withheld during synthesis, so the explanation must be supported by the image and its annotations. Pixel coordinates are converted to normalized permille coordinates in the order \texttt{[ymin,xmin,ymax,xmax]}, matching the representation used by ARGUS.

The filtering process removes examples whose structural validity cannot be assessed reliably. These include fictional anatomy without an applicable physical reference and images in which darkness, blur, or occlusion prevents inspection of the annotated structure. The purpose is to exclude ambiguous supervision, not to treat unusual appearance as an error. After filtering, 10K grounded examples are selected for the SFT cold-start. The remaining training images do not require synthesized explanations: their category and box annotations provide supervision for the subsequent reward-based training stage.

\begin{figure}[t]
    \centering
    \includegraphics[width=\linewidth]{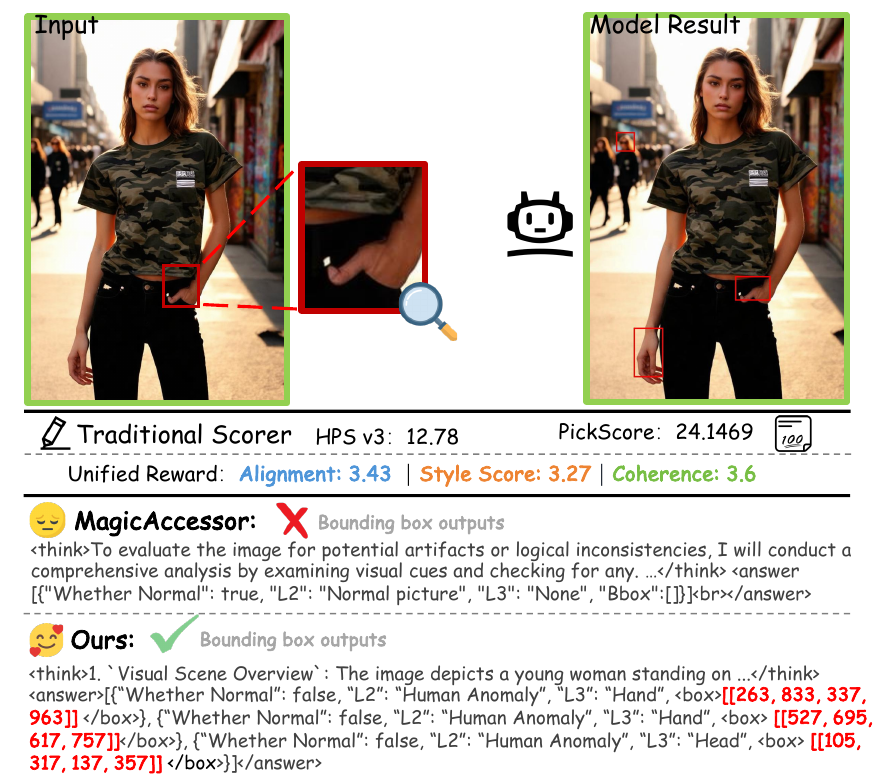}
    \caption{\textbf{Examples of diagnostic feedback.} Image-level scores summarize overall quality, while textual assessments can describe defects without locating them. ARGUS additionally predicts anomaly boxes, making its feedback usable for spatial reward assignment. The examples illustrate output formats rather than a general impossibility of grounding with other MLLMs.}
    \label{fig:supp_motivation}
\end{figure}

\subsection{Benchmark Composition}

The diagnostic benchmark contains 1,185 images with structural anomalies and 198 normal controls, totaling 1,383 images and 3,064 annotated boxes. It includes human, animal, interaction, and object-level defects, with a long-tailed distribution over finer subcategories. Human-hand errors are common, but the benchmark also includes less frequent failures involving animal anatomy and object structure.

The reported median box area is approximately $1{,}224~\mathrm{px}^{2}$. This is an absolute image-coordinate statistic, not a resolution-invariant measure of difficulty. Small defects and rare categories motivate reporting localization and category-level results alongside binary detection: strong aggregate detection alone does not establish reliable grounding across all structural categories.

\section{Implementation and Optimization Details}
\label{supp:experiments}

\subsection{Training Configuration and Computational Cost}

D2G is implemented in PyTorch with DeepSpeed ZeRO-2 and trained on 32 NVIDIA H800 GPUs with 80GB memory each. Stage I initializes ARGUS from Qwen3-VL-8B~\cite{qwen3vl}; Stage II updates the transformer backbone of FLUX.1-dev~\cite{flux2024}. Table~\ref{tab:supp_hyperparameters} lists the reported optimization settings.

\begin{table}[t]
    \caption{\textbf{Training configuration.} Gaussian smoothing is specified in latent-grid units. The Stage-I localization threshold and Stage-II confidence threshold serve different purposes.}
    \label{tab:supp_hyperparameters}
    \centering
    \small
    \setlength{\tabcolsep}{4pt}
    \begin{tabular}{@{}ll@{}}
        \toprule
        \textbf{Setting} & \textbf{Value} \\
        \midrule
        \multicolumn{2}{@{}l}{\textit{Stage I: SFT}} \\
        Training examples & 10K \\
        Global batch size & 128 \\
        Training epochs & 2 \\
        Peak learning rate & $2\times10^{-5}$ \\
        Learning-rate schedule & Cosine \\
        \midrule
        \multicolumn{2}{@{}l}{\textit{Stage I: Region-Aware GRPO}} \\
        Rollout group size & 8 \\
        Learning rate & $1\times10^{-6}$ \\
        KL coefficient & 0.01 \\
        IoU consistency threshold & 0.5 \\
        \midrule
        \multicolumn{2}{@{}l}{\textit{Stage II: Patch-level Flow-GRPO}} \\
        Rollout group size & 4 \\
        Learning rate & $5\times10^{-6}$ \\
        Training iterations & 1,000 \\
        Critical timestep & 0.3 \\
        Gaussian smoothing scale & 1.5 latent pixels \\
        Confidence threshold & 0.3 \\
        Patch-surrogate scale & $3/2$ \\
        \bottomrule
    \end{tabular}
\end{table}

The reported Stage-I training time is approximately 68 hours: 4 hours for the SFT cold-start and 64 hours for Region-Aware GRPO. A 1,000-iteration Stage-II run takes approximately 18 hours, and Stage II was evaluated with three random seeds. These are wall-clock measurements for the reported implementation, not evidence that TPSS alone accounts for the training cost.

An ARGUS diagnostic query was reported to take approximately 450 ms per image. This measures an additional diagnostic operation, not an unavoidable overhead of the aligned generator. ARGUS scores rollouts during Stage-II training; generating an image with the resulting model does not inherently require another ARGUS query. The timing should therefore be interpreted separately from generator inference latency.

% AUTHOR CHECK: Confirm whether the 18-hour timing is per seed. Document image resolution, sampling steps, precision, optimizer settings, gradient accumulation, and the batch size and decoding configuration used for the 450-ms diagnostic measurement. The supplied materials do not establish these settings or matched wall-clock budgets for all baselines.

\subsection{Matching, Spatial Rewards, and Confidence}

ARGUS uses Hungarian assignment to associate predicted and annotated boxes. GIoU supplies the matching and localization reward, whereas the reported consistency gate uses an IoU threshold of $0.5$. These operations should not be conflated. In particular, GIoU provides reward variation for non-overlapping boxes; ARGUS is optimized through sampled-response policy gradients, not by differentiating a coordinate-regression loss through its discrete box tokens.

% AUTHOR CHECK: Specify the implemented reward coefficients, normalization, treatment of unmatched predictions and annotations, and handling of empty box sets. A matched-box average alone does not define penalties for missed or additional boxes, or the reward for a normal image.

In Stage II, predicted boxes are mapped to the latent grid and smoothed with a Gaussian kernel of scale $1.5$. This softens abrupt spatial changes in the structural penalty. Unless the kernel is explicitly truncated, Gaussian smoothing extends beyond the original rectangle. Moreover, the image-level reward and shared generator parameters can affect regions outside the diagnostic mask. Spatially localized supervision therefore does not imply that all other pixels or parameter gradients remain unchanged.

The confidence safeguard reduces the structural penalty when the diagnosis is uncertain. Confidence is derived from the log-probabilities of generated bounding-box tokens. Denoting the resulting confidence score by $c$, the effective penalty can be written as
\begin{equation}
    \lambda(c)=\lambda_p g(c),
    \qquad 0\leq g(c)\leq 1,
    \label{eq:supp_confidence_decay}
\end{equation}
where the reported implementation applies exponential attenuation below a confidence threshold of $0.3$. This is a soft confidence-dependent weighting, not the hard indicator gate used for supervised localization.

The Stage-I IoU gate relies on annotated reference boxes. Generated Stage-II images do not provide such references, so a ground-truth GIoU test cannot be reused there without additional supervision. Token confidence is also only a heuristic indicator of reliability; the experiments do not establish that it is a calibrated probability of diagnostic correctness.

% AUTHOR CHECK: Replace g(c) with the exact implemented decay function and specify how box-token log-probabilities are aggregated and normalized. Do not substitute a hard threshold, a geometric mean, or an unreported decay coefficient. If Stage II also uses a box-consistency test, identify its reference boxes and distinguish it from ground-truth IoU/GIoU.

\subsection{TPSS and the Scope of Distributional Claims}
\label{supp:tpss_theory}

TPSS is defined at the rollout-group level. A group shares a prompt, an initial latent, and an early deterministic trajectory; its members branch through independent stochastic increments only after $t_{\mathrm{critical}}$. Independent initial latents would not provide the same spatial correspondence merely because their early integration steps were deterministic. The reported displacement analysis concerns correspondence within these shared-prefix groups.

For a stochastic sampling step, write the transition kernel as
\begin{equation}
    K_{\theta,t}(x'\mid x)
    =
    \mathcal{N}\!\left(x';\mu_{\theta,t}(x),\Sigma_t\right),
    \label{eq:supp_transition_kernel}
\end{equation}
where $\Sigma_t$ is positive definite and independent of $\theta$. The likelihood gradient is
\begin{equation}
    \nabla_\theta\log K_{\theta,t}(x'\mid x)
    =
    \bigl(\nabla_\theta\mu_{\theta,t}(x)\bigr)^{\mathsf T}
    \Sigma_t^{-1}\bigl(x'-\mu_{\theta,t}(x)\bigr).
    \label{eq:supp_kernel_gradient}
\end{equation}
The scaling of a velocity-based update therefore depends on the transition mean, covariance, timestep convention, and objective. We treat the $3/2$ coefficient in the patch surrogate as an empirical scaling choice, not a universal consequence of the SDE likelihood. The transition mean must also match the implemented sampler: adding noise to an ODE update is not, by itself, a marginal-preserving ODE-to-SDE conversion~\cite{liu2025flow,song2020score}.

% AUTHOR CHECK: Reconcile the main-paper rollout equation with the implemented SDE drift, noise schedule, and terminal-step handling. Equation~\ref{eq:supp_kernel_gradient} does not establish that a constant-scaled, spatially weighted surrogate is an unbiased gradient of the full rollout objective.

A conditional comparison is possible when two late-stage processes begin at the same branch state and use the same nondegenerate transition covariances. Let $P_z$ and $Q_z$ be their path distributions over $K$ steps, starting at $Y_0=z$, and let $\Delta\mu_k(y)=\mu_k^P(y)-\mu_k^Q(y)$. The chain rule for relative entropy and the Gaussian transition formula give
\begin{equation}
    \begin{aligned}
        D_{\mathrm{KL}}\!\left(P_z^{\mathrm{end}}\middle\|Q_z^{\mathrm{end}}\right)
        &\leq D_{\mathrm{KL}}(P_z\|Q_z) \\
        &=\frac{1}{2}\sum_{k=1}^{K}
        \mathbb{E}_{P_z}\!\left[
        \left\|\Delta\mu_k(Y_{k-1})\right\|_{\Sigma_k^{-1}}^2
        \right],
    \end{aligned}
    \label{eq:supp_conditional_kl}
\end{equation}
provided the expectations are finite, where $\|a\|_{\Sigma^{-1}}^2=a^{\mathsf T}\Sigma^{-1}a$. The inequality follows by marginalizing the path to its endpoint.

Equation~\ref{eq:supp_conditional_kl} is a conditional comparison between processes with matched initial states and covariances. It is not a bound between complete TPSS and full-SDE trajectories. Their early diffusion coefficients differ, and a deterministic prefix and a nondegenerate stochastic prefix can induce mutually singular path measures. Nor does shortening the stochastic interval alone provide a numerical bound: the mean discrepancies and inverse covariances must also be controlled. We therefore characterize TPSS through measured spatial drift, structural fidelity, and image-quality metrics rather than claiming a negligible global KL divergence or exact marginal preservation.

\section{Evaluation Protocols}
\label{supp:evaluation_protocols}

\subsection{Diagnostic Metrics and Forced-Box Baselines}

Binary detection treats images containing structural defects as positive and normal images as negative. Category-level precision, recall, and F1 are reported separately from box localization. Under the four-category reporting scheme, macro F1 is defined as the unweighted average of the four category F1 scores; it is not the harmonic mean of macro precision and macro recall.

% AUTHOR CHECK: The supplied main tables do not currently satisfy this definition. For ARGUS/Qwen3-VL, the four reported category F1 values average to 0.692275, whereas Table 1 reports 0.6735. Recompute the summaries from the original predictions or document the actual aggregation protocol; do not silently replace one reported result with the other. Also specify the mIoU denominator and treatment of unmatched boxes.

To separate output-format limitations from localization accuracy, GPT-4.1 and Gemini 3.1 Pro are additionally evaluated with structured CoT examples requiring normalized \texttt{[ymin,xmin,ymax,xmax]} boxes. Table~\ref{tab:supp_forced_bbox} reports the available results under this protocol. Both models produce spatial outputs, but their localization remains less accurate than ARGUS on the evaluated defects. Missing localization measurements for other evaluators should be marked as unreported, not interpreted as zero accuracy or an inability to emit coordinates.

\begin{table}[t]
    \caption{\textbf{Forced-box diagnostic evaluation.} Structured CoT prompting requires proprietary models to emit bounding boxes. Precision and recall refer to binary anomaly detection.}
    \label{tab:supp_forced_bbox}
    \centering
    \small
    \setlength{\tabcolsep}{3pt}
    \begin{tabular}{@{}lccc@{}}
        \toprule
        \textbf{Model} & \textbf{Prec.} & \textbf{Rec.} & \textbf{mIoU} \\
        \midrule
        GPT-4.1~\cite{GPT-4.1} & 0.6315 & 0.4055 & 0.2361 \\
        Gemini 3.1 Pro~\cite{Gemini-3-pro} & 0.5577 & \textbf{0.8214} & 0.2814 \\
        ARGUS & \textbf{0.8655} & 0.8012 & \textbf{0.7540} \\
        \bottomrule
    \end{tabular}
\end{table}

The forced-box experiment reports binary precision, recall, and mIoU. It does not separately establish that the original L2 or human-alignment scores remain unchanged under the new prompt. Human-alignment accuracy for the diagnostic model, following~\cite{wang2026thinking}, should also be distinguished from the generated-image preference study described below.

\subsection{Structural Fidelity and Subject Verification}

Generative evaluation uses the separate 800-prompt Achilles set. Prompts require the target subject or challenging body part to be visible, reducing the opportunity to avoid evaluation by hiding or omitting it. Let $V_i$ indicate that output $i$ satisfies subject verification, and let $D_i$ indicate that a verified output contains a structural defect. Counting visibility failures as unsuccessful outputs gives
\begin{equation}
    S_{\mathrm{struct}}
    =
    \frac{100}{N}\sum_{i=1}^{N}V_i(1-D_i).
    \label{eq:supp_structural_score}
\end{equation}
Equivalently, unsuccessful outputs comprise both subject-verification failures and defective verified images. Removing visibility failures from the denominator would instead measure quality conditional on successful subject rendering and would not penalize generative evasion.

Overall fidelity aggregates evaluated outputs; it should not be assumed to equal the unweighted mean of the four category columns unless the category sizes and aggregation procedure justify that equality. Global image quality is assessed separately with FID-30K~\cite{FID} and CLIP Score~\cite{hessel2021clipscore} on 30,000 MS-COCO prompts.

% AUTHOR CHECK: Confirm that all reported fidelity values use the same visibility-failure convention. Record category sample counts, output counts per prompt, seed aggregation, the subject-verification model, and invalid-output handling. Specify the COCO split and reference images, FID implementation, and CLIP checkpoint.

\subsection{Independent Assessment and Local Baselines}

ARGUS supplies the training reward, so its evaluation alone cannot establish that the generator improves beyond that evaluator. We additionally assess outputs using HumanRefiner~\cite{HumanRefiner}, zero-shot GPT-4.1, and human preference. These complementary measurements reduce dependence on ARGUS but do not eliminate all evaluator bias.

HumanRefiner provides a human-centric structural assessment. Its anomaly-free rate is therefore not numerically interchangeable with the cross-category Achilles fidelity score. Under this external protocol, D2G obtains $74.8\%$, compared with $42.2\%$ for the base model and $44.5\%$ for Global Flow-GRPO. The complete comparison is reported in Table~\ref{tab:generative_fidelity}.

The local baselines include masked inpainting and an adaptation of LocalDPO~\cite{localdpo}. The latter replaces random masks in its masked-preference mechanism with ARGUS semantic masks, giving the baseline access to detector-derived spatial information. Masked inpainting and adapted-LocalDPO obtain HumanRefiner anomaly-free rates of $68.2\%$ and $60.5\%$, respectively. The adaptation is reported as \emph{adapted-LocalDPO}, rather than as a reproduction of its original task setting.

% AUTHOR CHECK: Document the HumanRefiner evaluation subset and its size. Specify the inpainting model and mask construction, the preference-pair construction for adapted-LocalDPO, and the training and sampling budgets. Shared diagnostic masks do not by themselves establish equal computational budgets.

For evaluation beyond Achilles prompts, we use 1,000 unseen human-centric MS-COCO prompts. HumanRefiner reports $71.5\%$ anomaly-free outputs for D2G and $41.8\%$ for the base model. These scores come from a different prompt set and should remain separate from the $74.8\%/42.2\%$ comparison above.

\subsection{Human Preference Study}

The double-blind study uses 300 randomly sampled image pairs from Achilles-Benchmark and 15 evaluators with backgrounds in digital arts, computer vision, and photography. Each comparison presents the prompt and two outputs, one from the base FLUX.1-dev model and one from an aligned model. Model identities are hidden and left/right positions are randomized.

Evaluators choose a win, tie, or loss using structural rationality as the primary criterion. They inspect anatomical and physical inconsistencies, including malformed limbs, incorrect finger counts, fused objects, and implausible interactions. When structural quality is comparable, global appearance provides the secondary criterion, including background texture, lighting, color, and overall coherence. Each pair receives three judgments, which are aggregated by majority consensus. The reported inter-rater agreement is Fleiss' $\kappa=0.78$.

D2G receives a reported human preference score of $82.5\%$. This is a pairwise quality assessment, not the percentage of images with no structural defects. It also differs from the $86.7\%$ human-alignment accuracy reported for ARGUS as a diagnostic evaluator.

% AUTHOR CHECK: Recover the win/tie/loss counts, the resolution of three-way vote disagreements, and the exact convention used for the reported percentages. The source material permits ties but does not specify tie weighting or the denominator after aggregation. Do not infer these choices from the decimal percentages.

\subsection{Zero-Shot GPT-4.1 Assessment}

GPT-4.1~\cite{GPT-4.1} is used as an additional judge without task-specific fine-tuning. It receives the prompt and a pair of generated images and returns a brief explanation and a discrete preference. The evaluation uses 300 test comparisons and the following prompt.

\begin{tcolorbox}[colback=gray!5,colframe=gray!50,title=GPT-4.1 Zero-Shot Judge Prompt]
\small
\textbf{System Prompt:} You are an impartial and expert judge for generative AI visual quality. Your task is to evaluate two AI-generated images (Image A and Image B) based on a given text prompt.

\textbf{Evaluation Criteria:}

1. \textbf{Geometric and Structural Fidelity:} Inspect the semantic subjects carefully. Which image possesses better physical and anatomical logic? Look for severe artifacts such as malformed limbs, incorrect number of fingers, or fused objects.

2. \textbf{Holistic Aesthetic Quality:} Look at the background, lighting, and textures. Which image preserves a higher level of photorealism and global coherence without looking artificially blurred, over-smoothed, or degraded?

\textbf{User Input:}

Text Prompt: \texttt{\{t2i\_prompt\}}

[Image A] [Image B]

\textbf{Output Format:} You must output a JSON object with two keys: ``Reasoning'' (a brief explanation of your comparative analysis based on the criteria) and ``Choice'' (strictly output ``A'', ``B'', or ``Tie'').
\end{tcolorbox}

The reported GPT-4.1 preference score for D2G is $80.2\%$. Agreement with the human study provides evidence that the improvement is not confined to ARGUS scoring. However, an independent model can still share perceptual or stylistic biases with the training evaluator; this assessment is not an unbiased ground truth.

% AUTHOR CHECK: Record the API model identifier, evaluation date, decoding settings, image ordering, invalid-response handling, and tie convention used to obtain the reported score. Do not describe order-swapped evaluation or repeated judging unless those runs were actually performed.

\section{Additional Results and Analysis}
\label{supp:additional_results}

\subsection{Transfer to SD3.5-Medium}

We apply Stage II to Stable Diffusion 3.5-Medium~\cite{sd3} to examine whether the improvement extends beyond FLUX.1-dev. Table~\ref{tab:supp_sd35} preserves the complete reported comparison. Global Flow-GRPO increases overall structural fidelity from $54.82$ to $58.20$, while FID rises from $20.15$ to $26.40$. Patch-level Flow-GRPO reaches $81.50$ overall fidelity and $76.45$ human fidelity, with FID close to the base-model value. This supports transfer across the two evaluated generators, rather than a claim of architecture-independent performance.

\begin{table}[t]
    \caption{\textbf{Stage-II alignment on SD3.5-Medium.} Fidelity is evaluated on Achilles-Benchmark. Higher fidelity and lower FID are preferable.}
    \label{tab:supp_sd35}
    \centering
    \small
    \setlength{\tabcolsep}{3pt}
    \begin{tabular}{@{}lccc@{}}
        \toprule
        \textbf{Alignment} & \shortstack{\textbf{Human}\\\textbf{Fidelity}} & \shortstack{\textbf{Overall}\\\textbf{Fidelity}} & \textbf{FID-30K} \\
        \midrule
        Base SD3.5-M & 39.33 & 54.82 & 20.15 \\
        Global Flow-GRPO & 42.10 & 58.20 & 26.40 \\
        Patch-level Flow-GRPO & \textbf{76.45} & \textbf{81.50} & \textbf{20.10} \\
        \bottomrule
    \end{tabular}
\end{table}

\subsection{Critical Timestep and Spatial Correspondence}
\label{supp:tpss_sensitivity}

\begin{figure}[t]
    \centering
    \includegraphics[width=0.95\linewidth]{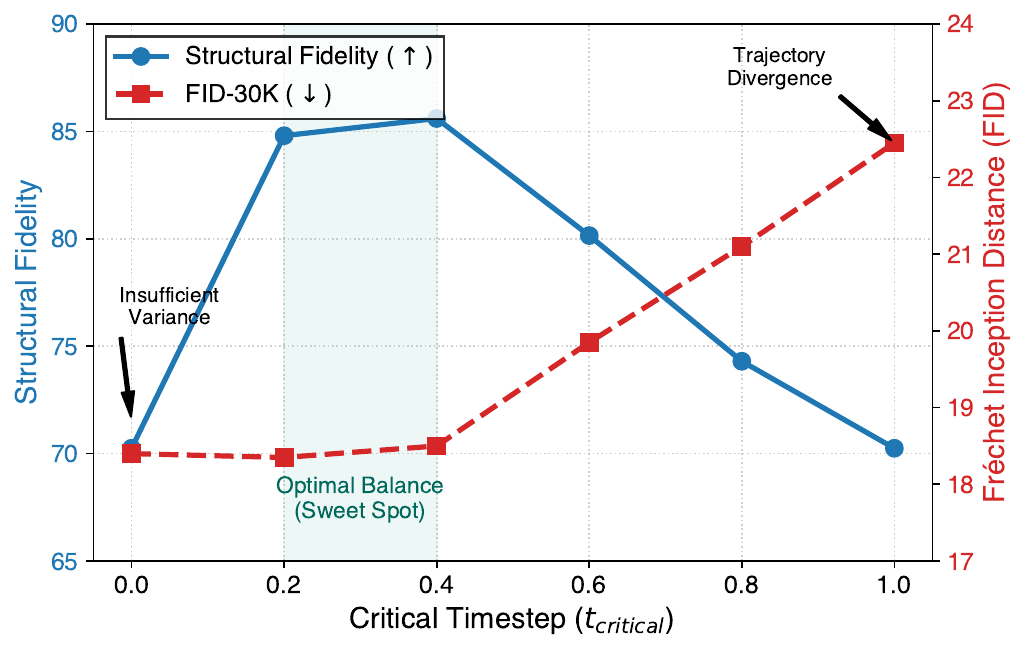}
    \caption{\textbf{Sensitivity to the TPSS branching timestep.} The original sweep reports structural fidelity and FID at different values of $t_{\mathrm{critical}}$. Intermediate thresholds provide a better trade-off than nearly deterministic sampling or early stochastic exploration.}
    \label{fig:supp_t_critical}
\end{figure}

The original sensitivity sweep evaluates $t_{\mathrm{critical}}\in\{0.0,0.2,0.4,0.6,0.8,1.0\}$, as shown in Figure~\ref{fig:supp_t_critical}. An additional analysis evaluates $\{0.1,0.2,0.3,0.4,0.6\}$. Both identify a favorable trade-off within $[0.2,0.4]$. These evaluations should not be merged into an interpolated curve without the underlying per-setting measurements.

Small thresholds leave little stochastic variation for group-relative optimization. With a shared initial latent, fully deterministic sampling also removes rollout diversity unless another source of randomness is introduced. Larger thresholds permit earlier stochastic perturbations and increase the opportunity for semantic regions to move across the group. The observed trade-off supports the selected default of $0.3$; it does not establish that structural geometry is determined exclusively at late timesteps.

At $t_{\mathrm{critical}}=0.3$, the reported mean displacement of matched patches is $1.8\pm0.6$ pixels on a $1024\times1024$ canvas, compared with more than 15 pixels under global SDE sampling. This describes the evaluated rollout groups, not a uniform bound on spatial movement. Consistent with this observation, the no-TPSS ablation reduces structural fidelity from $85.60$ to $70.25$ and increases FID from $18.35$ to $22.45$.

% AUTHOR CHECK: Specify the matching algorithm, displacement definition, number of prompts and matched regions, and meaning of the reported +/- 0.6. Do not label it as a standard deviation, standard error, or confidence interval without the original measurement record. Confirm that the retained figure uses the stated sweep and the same evaluation configuration as its caption.

\subsection{Grounded Supervision and Diagnostic Robustness}
\label{supp:robustness}

Table~\ref{tab:supp_pipeline_ablation} reports the additional controls for Reverse-CoT and Confidence-Decay. Replacing grounded Reverse-CoT synthesis with open-ended VQA reduces diagnostic F1 to $0.751$ and mIoU to $0.615$. Removing Confidence-Decay retains much of the structural fidelity but increases FID to $21.40$. These results complement the matching, consistency-gate, sampling, TPSS, and soft-mask ablations in Table~\ref{tab:ablation}.

\begin{table}[t]
    \caption{\textbf{Additional pipeline ablations.} Reverse-CoT is evaluated through Stage-I diagnostics; Confidence-Decay is evaluated through Stage-II generation. Precision is retained as reported in the corresponding experiments.}
    \label{tab:supp_pipeline_ablation}
    \centering
    \small
    \setlength{\tabcolsep}{2.5pt}
    \begin{tabular}{@{}lcccc@{}}
        \toprule
        \textbf{Variant} & \shortstack{\textbf{Diag.}\\\textbf{F1}} & \textbf{mIoU} & \textbf{Fidelity} & \textbf{FID} \\
        \midrule
        Full D2G & \textbf{0.8321} & \textbf{0.7540} & \textbf{85.60} & \textbf{18.35} \\
        w/o Reverse-CoT & 0.751 & 0.615 & -- & -- \\
        w/o Confidence-Decay & -- & -- & 84.15 & 21.40 \\
        \bottomrule
    \end{tabular}
\end{table}

An additional stress test injects 10\% false-positive anomaly boxes during Stage-II training, targeting otherwise valid hands and objects. The available assessment of this experiment is qualitative; no separate aggregate scores or uncertainty estimates were reported. It therefore does not establish a quantified tolerance to a 10\% diagnostic error rate. The Confidence-Decay ablation supplies the quantitative evidence for the safeguard, while systematic robustness to different error rates remains unmeasured.

The soft-mask ablation provides a separate control. Replacing Gaussian smoothing with a hard mask reduces fidelity to $83.90$ and increases FID to $19.75$. This is consistent with a benefit from smoother spatial weighting, but does not establish that every quality difference is caused by boundary artifacts alone.

\subsection{Limitations and Interpretation}

D2G depends on the diagnostic model's coverage and reliability. Severe occlusion, ambiguous structure, and categories excluded during data filtering remain difficult cases; confidence weighting reduces their influence without guaranteeing correct rejection. The external evaluation also has limits: HumanRefiner is human-centric, and human and MLLM preferences combine structural and appearance judgments.

Finally, spatial concentration of the training signal is not a guarantee of unchanged surrounding content. TPSS reduces measured rollout drift but does not provide exact correspondence or a verified global distributional bound. The reported FID values support preservation under the evaluated settings; small differences such as $18.42$ versus $18.35$ should not be interpreted as statistically significant improvements without uncertainty estimates.

\end{document}